\documentclass[runningheads]{llncs}
\usepackage{hyperref}
\usepackage{graphicx}
\usepackage{xcolor}
\usepackage{siunitx}
\usepackage{amsmath}
\usepackage{amssymb}
\usepackage{subcaption}
\usepackage{algorithm, algpseudocode}
\usepackage{multirow}
\usepackage{tikz}
\usepackage{booktabs}

\newcommand{\marta}[1]{\textcolor{red}{Marta: #1}}

\begin{document}


\title{The Quest of Finding the Antidote to Sparse Double Descent}
\author{Victor Qu\'etu \and
Marta Milovanović}


\institute{LTCI, T\'el\'ecom Paris, Institut Polytechnique de Paris}

\maketitle

\begin{abstract}
In energy-efficient schemes, finding the optimal size of deep learning models is very important and has a broad impact. Meanwhile, recent studies have reported an unexpected phenomenon, the sparse double descent: as the model’s sparsity increases, the performance first worsens, then improves, and finally deteriorates. Such a non-monotonic behavior raises serious questions about the optimal model’s size to maintain high performance: the model needs to be sufficiently over-parametrized, but having too many parameters wastes training resources. 

In this paper, we aim to find the best trade-off efficiently. More precisely, we tackle the occurrence of the sparse double descent and present some solutions to avoid it. Firstly, we show that a simple $\ell_2$ regularization method can help to mitigate this phenomenon but sacrifices the performance/sparsity compromise. 
To overcome this problem, we then introduce a learning scheme in which distilling knowledge regularizes the student model. Supported by experimental results achieved using typical image classification setups, we show that this approach leads to the avoidance of such a phenomenon.

\end{abstract}

\section{Introduction}
\label{sec:Intro}
\let\svthefootnote\thefootnote
\newcommand\freefootnote[1]{%
  \let\thefootnote\relax%
  \footnotetext{#1}%
  \let\thefootnote\svthefootnote%
}

The\freefootnote{This paper has been accepted for publication at the Workshop on
Simplification, Compression, Efficiency and Frugality for Artificial intelligence (SCEFA) in conjunction with ECML PKDD 2023.} field of computer vision has undergone a remarkable transformation with the advent of deep neural networks (DNNs). These models possess the ability to learn high-level representations of features from raw input data~\cite{dosovitskiy2021an}. Compared to conventional machine learning algorithms, DNNs have demonstrated superior performance across various visual recognition tasks. Notably, they have achieved state-of-the-art results in challenging areas such as segmentation~\cite{chaudhry2022lung}. Moreover, DNNs have excelled in image classification~\cite{barbano2022two}, as well as in object detection~\cite{mazzeo2022image}. DNNs have the capacity to process large amounts of data, enabling them to capture intricate patterns and details. By training on extensive datasets, DNNs can generalize their learned knowledge to effectively recognize and interpret novel, previously unseen examples. This ability to generalize contributes to their robustness and adaptability in real-world scenarios.


However, a significant disadvantage of DNNs is that they are prone to overfitting; as a result, learning procedures need to be created to counteract it. The most direct solution would be to increase the dataset size, as deep learning techniques are well-known for their data hunger. 
Since they usually optimize some objective function through gradient descent, having a larger training set helps the optimization process to find the most appropriate set of features, resulting in high performance on unseen data. But, such an approach presents two shortcomings: it requires enormous computational power for training and large annotated datasets. While addressing the first drawback remains an actual research topic~\cite{frankle2018the,bragagnolo2022update}, the second is broadly explored with approaches like transfer learning~\cite{zhuang2020comprehensive} or self-supervised learning~\cite{ravanelli2020multi}. 

\begin{figure}[t]
\centering
    \includegraphics[width=0.8\linewidth]{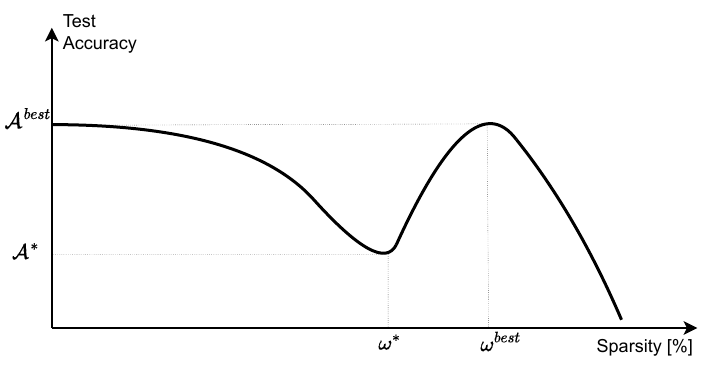}
    \caption{The Sparse Double Descent phenomenon: as the model’s sparsity increases, the performance first worsens, then improves, and finally deteriorates.}
    \label{fig:SDD_scheme}
\end{figure}

In reality, large datasets are typically unavailable: in the context of \emph{frugal AI}, techniques requiring a small amount of data need to be used, arising research questions on the enlargement of available datasets or the transfer of knowledge from similar tasks. However, it brings also into question the optimal dimension of the deep learning model to be trained. 
As opposed to what the bias-variance trade-off suggests, the \emph{double descent} (DD) phenomenon can be observed in a very over-parameterized network~\cite{belkin2019reconciling}: given some set of parameters $\boldsymbol{w}$ for the model with accuracy value $\mathcal{A}$, adding parameters will first improve the performance until a local maximum $\mathcal{A}^{best}$ beyond which, adding even more parameters, will worsen the performance until a local minimum $\mathcal{A}^{*}$, before going back increasing.
Regularly reported in the literature~\cite{spigler2019jamming,geiger2019jamming}, this phenomenon 
raises the challenge of finding the best set of parameters, avoiding entering an over-parametrized or under-parameterized regime.
Similar to this, a \emph{sparse double descent} (SDD) phenomenon is exhibited when the model shifts via parameter pruning from an over-parametrized towards a sparser regime~\cite{SparseDoubleDescent}, displayed in Fig.~\ref{fig:SDD_scheme}. 
Finding $\boldsymbol{w}^{best}$ is a problem requiring a lot of computation, or extremely over-sizing the model, which are the two possible solutions to address this challenge.
Since neither of these two approaches are suitable for a frugal setup, we are investigating if a solution to address this problem can be found.

Our work is structured as follows. First, we highlight the occurrence of the sparse double descent phenomenon in traditional image classification setups (Sec.~\ref{sec:Occurrence of Sparse Double Descent}).
Then, we show that a standard regularization method like $\ell_2$ regularization can help to eschew sparse double descent and we underline its limits (Sec.~\ref{sec:wd}).
Moreover, to overcome these limits, we introduce a learning scheme in which a student model is regularized by distilling knowledge from a sparse teacher in its best validation accuracy region (or even with a dense teacher) observing that the student is shunning such a phenomenon (Sec.~\ref{sec:kd}). Finally, we discuss the environmental impact and the potential future directions of our work (Sec.~\ref{sec:Environmental}).
\section{Related Works}
\label{sec:SOTA}

\textbf{Noisy data in the real world.} 
In the real world, the acquisition of data or the labeling process can typically lead to a noisy data collection~\cite{gupta2019dealing}.
Numerous works have addressed the issue of annotation noise and suggested ways to avoid learning the incorrect feature sets. For instance, \cite{li2017learning} proposes a unified distillation framework that uses the knowledge gained from a small, clean dataset and a semantic knowledge graph to correct noisy labels. The advantages of noise on the neurological system are used as inspiration in other works to discover solutions: in~\cite{arani2021noise}, it has been demonstrated how adding constructive noise to the collaborative learning framework at various levels makes it possible to train the model effectively and extract desirable traits from the student model. 
As a single image might be classified under multiple categories, distinct samples might have label noise of varied intensities. The normalized knowledge distillation feature, which introduces the sample-specific correction factor to replace the temperature, was proposed by~\cite{xu2020feature}. By using a teacher-student strategy, \cite{kaiser2022blind} was able to estimate the noise in the training data using Otsu's technique and determine the border between overfitting and good generalization. Other studies concentrate more on the accuracy of the label's prediction. For example, \cite{sau2016deep} presented a straightforward technique that, by introducing noise and perturbing the teacher's logit outputs, improves student learning and yields results that are more similar to the teacher network.
Because it is present in the loss layer in this configuration, the noise replicates a multi-teacher environment and has the effect of a regularizer. A common method is to manually introduce noise into some well-annotated, well-known datasets, such as MNIST and CIFAR-10/100, in order to simulate noise in the labels~\cite{nakkiran2021deep,SparseDoubleDescent}. A similar approach is used in AI security studies, where noise is parametrically inserted to assess the model's resilience to attacks. This approach is known as ``adversarial learning''~\cite{9013065} because these attacks offer adversarial interpretations of the data and evaluate the performance of the model.

\textbf{From double descent to sparse double descent in classification tasks.}
Given the existence of labeling noise, the threat posed by DD is real. Multiple machine learning models, including decision trees, random features~\cite{meng2022multiple}, linear regression~\cite{muthukumar2020harmless}, and deep neural networks~\cite{yilmaz2022regularization}, have already reported the existence of the DD phenomenon. For classification tasks, standard deep neural networks, such as the ResNet architecture, trained on image classification datasets consistently follow a double descent curve both when label noise is injected (CIFAR-10), and in some cases, even without any label noise injection (CIFAR-100)~\cite{yilmaz2022regularization}. The study~\cite{nakkiran2021deep} demonstrates that double descent happens as a function of both the number of training epochs and as the  model size when the model width is increased. Indeed, the double descent phenomenon has been thoroughly investigated within the context of over-parametrization~\cite{nakkiran2021deep,chang2021provable}. As opposed to the DD phenomenon, the SDD phenomenon occurs when a dense model is unstructurely pruned, in the transition from the complex model to the sparse, pruned model (as illustrated in Fig.~\ref{fig:SDD_scheme})~\cite{SparseDoubleDescent}. Given that SDD renders many criteria, such as when to stop the pruning, uncertain, it has consequences for model selection, regularization methods, and understanding the behavior of complicated models in high-dimensional contexts.

\textbf{Frugal AI and sparse double descent.} Frugality involves working with limited resources and it manifests in different ways~\cite{evchenko2021frugal}. One may consider input frugality, which focuses on the costs associated with data and it may involve a reduced amount of training data or fewer features than in the non-frugal scenario. Another type is learning process frugality, which emphasizes the costs associated with the learning process itself, including computational and memory resources. Finally, model frugality focuses on the costs associated with storing or using machine learning models (e.g. classifiers or regression models), driven by resource constraints like low memory or processing power. Frugal models for supervised learning may need less memory and generate predictions with less computing power than what is necessary for optimal prediction quality.

To reduce the computational footprint of models at training and inference, the training of sparse neural networks has been studied for resource-constrained
settings~\cite{schwarz2021powerpropagation}.
Indeed, sparse double descent is closely linked to frugality. Such a phenomenon typically occurs while training sparse models and it raises the challenge of finding the best set of parameters during model pruning.

In the next section, we will present and show the occurrence of the SDD phenomenon.

\section{Occurrence of Sparse Double Descent}
\label{sec:Occurrence of Sparse Double Descent}


\textbf{Exposing sparse double descent.} We present an algorithm to expose the sparse double descent phenomenon in Alg.~\ref{Algo}.
First, the dense model, represented by its parameters $\boldsymbol{w}$, is trained on the learning task $\Xi$, eventually with a $\ell_2$ regularization weighted by $\lambda$ (line~\ref{line:dense}). Then, the model is pruned (line~\ref{line:prune}), using an unstructured pruning method called magnitude-based pruning popularized by~\cite{han2015learning} in which a given amount of weights $\zeta^{iter}$, below some specific threshold, are pruned. 
At each pruning iteration, a fixed $\zeta^{\text{iter}}$ amount of weights from the model is removed. 
Magnitude-based pruning is used because of its competitiveness in terms of both effectiveness and computational simplicity~\cite{Gale_Magnitude} despite the existence of more complex pruning methods.
Since after pruning the performance of the model is affected, the model is retrained using the same original policy (line~\ref{line:wd}).
Recent works have shown that this approach leads to the best performance at the highest sparsities~\cite{quetu2023dodging}. Once the sparsity's model reaches $\zeta^{end}$, the pruning process ends. 

\begin{algorithm}[t]
\caption{Iterative magnitude pruning algorithm to expose SDD.}
\label{Algo}
\begin{algorithmic}[1]
\Procedure{Sketch ($\boldsymbol{w}^{init}$, $\Xi$, $\lambda$, $\zeta^{iter}$,$\zeta^{end}$)}{}
\State $\boldsymbol{w} \gets$ Train($\boldsymbol{w}^{init}$, $\Xi$, $\lambda$)\label{line:dense}
\While{Sparsity($\boldsymbol{w}, \boldsymbol{w}^{init}$) $< \zeta^{end}$}\label{line:endcond}
\State $\boldsymbol{w} \gets$ Prune($\boldsymbol{w}$, $\zeta^{iter}$) \label{line:prune}
\State $\boldsymbol{w} \gets$ Train($\boldsymbol{w}$,$\Xi$, $\lambda$)\label{line:wd} 
\EndWhile
\EndProcedure
\end{algorithmic}
\end{algorithm}

\textbf{Setup.} The same approach as He~et~al.~\cite{SparseDoubleDescent} is followed for the experimental setup. The first model we train is ResNet-18, trained on CIFAR-10 \& CIFAR-100, for 160 epochs, optimized with SGD, having momentum 0.9, a learning rate of 0.1 decayed by a factor 0.1 at milestones 80 and 120, batch size 128 and $\lambda$ equal to 1e-4. The second model is ViT, with 4 patches, 8 heads, and 512 embedding dimensions, trained on CIFAR-10 and CIFAR-100 for 200 epochs, optimized with SGD, having a learning rate of 1e-4 with a cosine annealing schedule and $\lambda$ equal to 0.03.
For each dataset, a percentage $\varepsilon$ of symmetric, noisy labels is introduced: the labels of a given proportion of training samples are flipped to one of the other class labels, selected with equal probability~\cite{Noisy_labels}. 
In our experiments, we set $\varepsilon \in \{10\%, 20\%, 50\%\}$. Moreover, we also conduct experiments without adding synthetic noise, as it has clean structures which greatly enable statistical analyses but often fail to model real-world noise patterns. Towards this end, we carry out experiments also on CIFAR-100N, a dataset presented by~\cite{wei2022learning}, which is formed by the CIFAR-100 training dataset equipped with human-annotated real-world noisy labels collected from Amazon Mechanical Turk. Thus, we use the same architectures and learning policies presented above. 
In all experiments, we use $\zeta^{\text{iter}}$ to 20\% and $\zeta^{\text{end}}>$ 99.8\%.

\begin{figure*}[t]
    \begin{subfigure}{0.5\textwidth}
        \includegraphics[width=\textwidth]{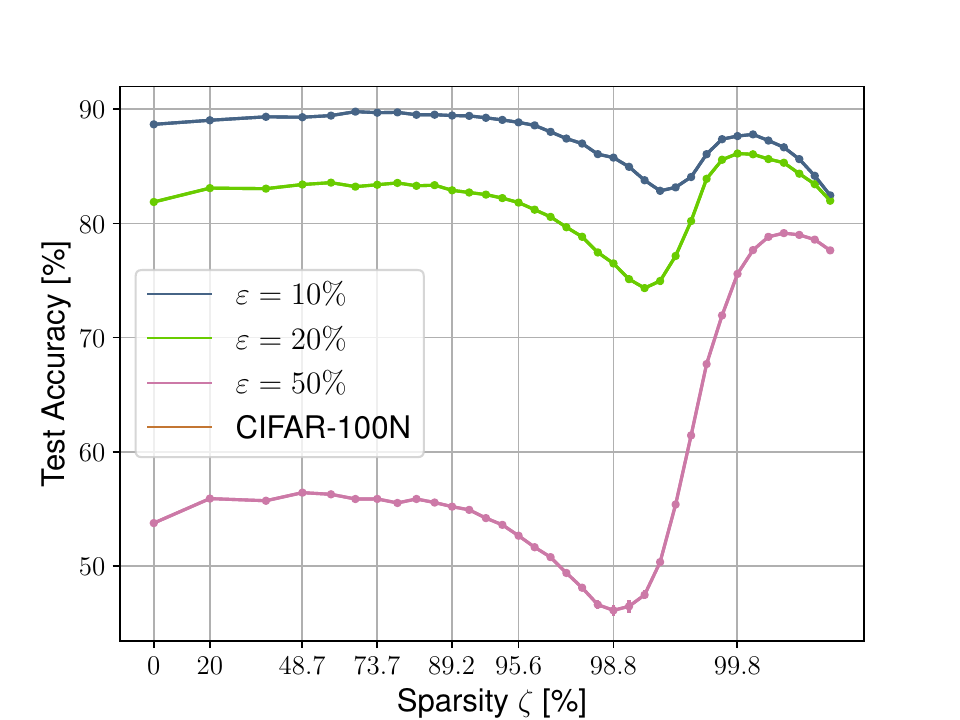}
        \caption{~}
        \label{fig:C10-R18}
    \end{subfigure}
    \begin{subfigure}{0.5\textwidth}
        \includegraphics[width=\textwidth]{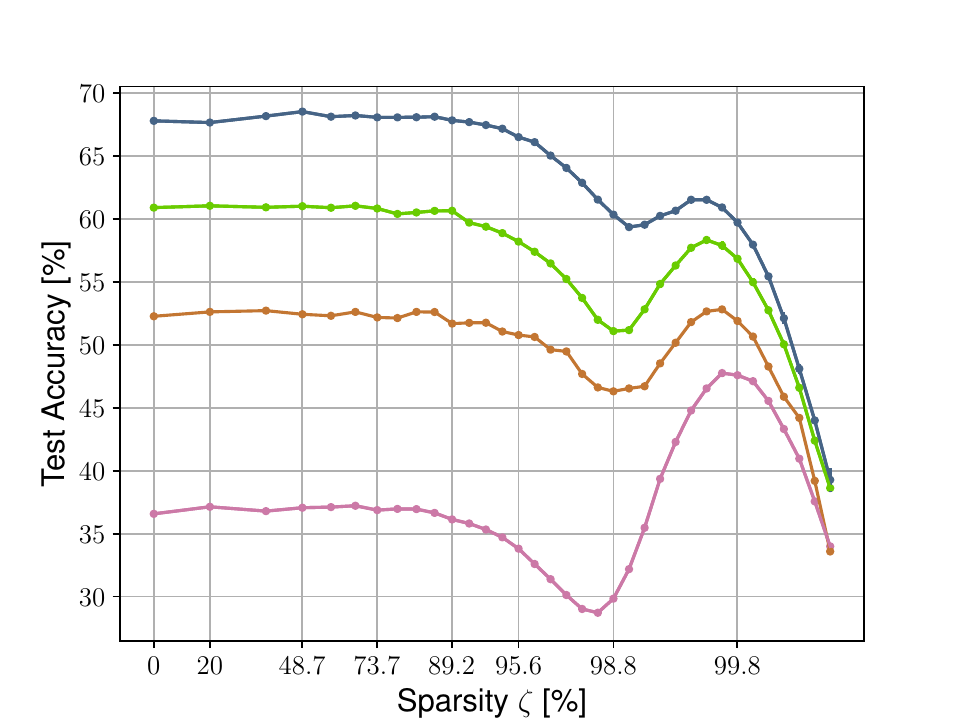}
        \caption{~}
        \label{fig:C100-R18}
    \end{subfigure}
    \label{fig:CIFAR-R18}
    \begin{subfigure}{0.5\textwidth}
        \includegraphics[width=\textwidth]{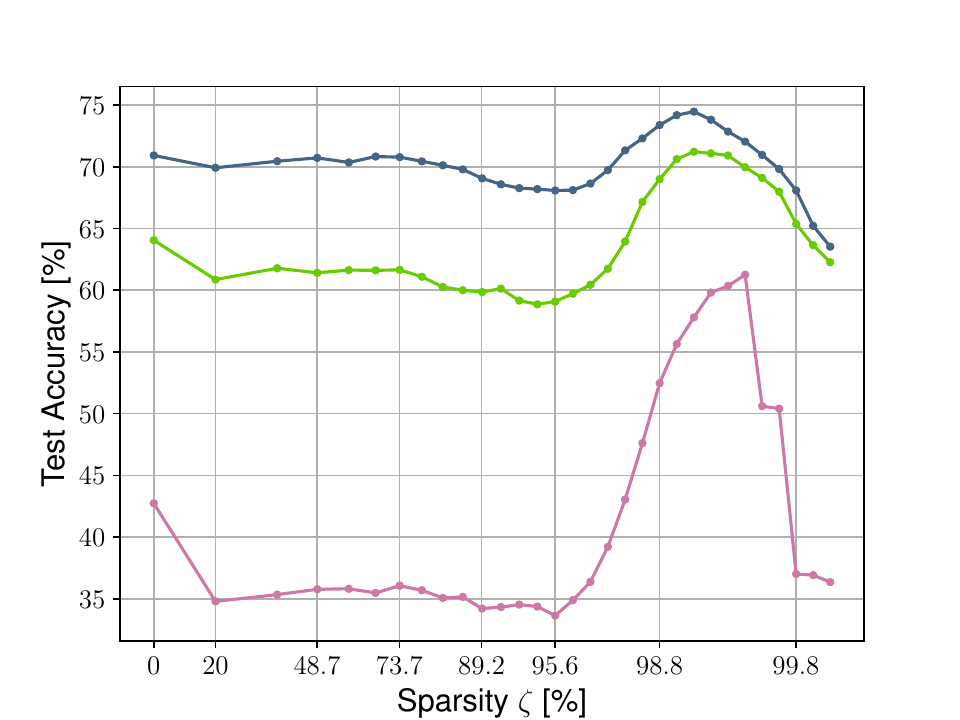}
        \caption{~}
        \label{fig:C10-ViT}
    \end{subfigure}
    \begin{subfigure}{0.5\textwidth}
        \includegraphics[width=\textwidth]{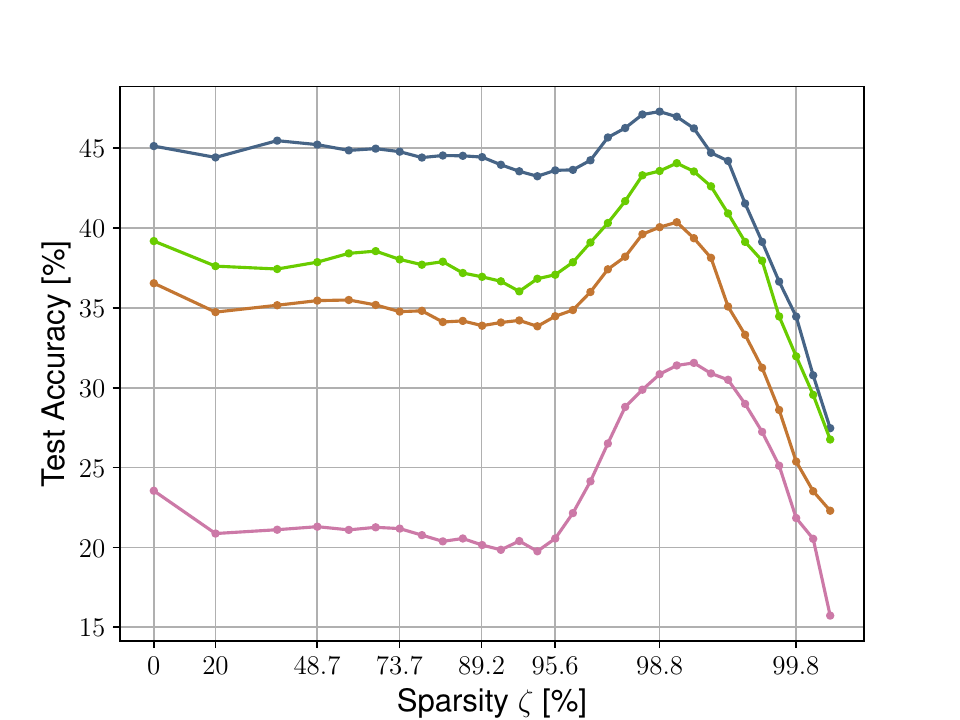}
        \caption{~}
        \label{fig:C100-ViT}
    \end{subfigure}
    \caption{Test accuracy of ResNet-18 on CIFAR-10~(a) and CIFAR-100~(b) ($\lambda=\num{1e-4}$), and on ViT on CIFAR-10~(c) and CIFAR-100~(d) ($\lambda=0.03$).}
    \label{fig:CIFAR-ViT-ResNet}
    \vspace{-10pt}
\end{figure*}

\textbf{Occurrence of sparse double descent.} 
Fig.~\ref{fig:CIFAR-ViT-ResNet} displays the results of ResNet-18 and ViT, on CIFAR-10 and CIFAR-100. As in He~et~al.~\cite{SparseDoubleDescent} work, the double descent consists of 4 phases. First, at low sparsities, the network is overparameterized, thus pruned network can still reach similar accuracy to the dense model. The second phase is a phase near the ``interpolation threshold'', where training accuracy is going to drop, and test accuracy is about to first decrease and then increase as sparsity grows. The third phase is located at high sparsities, where test accuracy is rising. The final phase happens when both training and test accuracy drop significantly.
For all the investigated noise rates $\varepsilon$, whether on CIFAR-10 or CIFAR-100, the sparse double descent phenomenon occurs both for ResNet and ViT. We observe a similar phenomenon in the simulated $\varepsilon$ of the human-annotated CIFAR-100N.

Looking at the curves in Fig.~\ref{fig:CIFAR-ViT-ResNet}, it is clear that it is difficult to define the traditional early stopping criteria. Indeed, they may make the mistake of stopping the algorithm when the model reaches a minimum performance in the second phase, whereas it increases again during the next phases.
Avoiding the sparse double descent phenomenon would allow us to use the traditional stopping criteria. Indeed, if the performance becomes monotonic, we can stop the algorithm when the performance starts deteriorating, knowing that it will never improve again after that point, enabling us to save computational training costs and find easily the best performance/sparsity trade-off.

Toward this end, we present in the next section an approach using $\ell_2$ regularization to avoid SDD. 
\section{$\ell_2$ Regularization Helps to Lessen the Sparse Double Descent}
\label{sec:wd}
\textbf{The positive contribution of $\ell_2$ regularization.} Recently, it has been shown that, for certain linear regression models with isotropic data distribution, optimally-tuned $\ell_2$ regularization can achieve monotonic test performance as either the sample size or the model size is grown. Nakkiran~et~al.~\cite{nakkiran2021optimal} demonstrated it analytically and established that optimally-tuned $\ell_2$ regularization can mitigate double descent for general models, including neural networks like Convolutional Neural Networks.
However, for classification tasks, this problem is not easily alleviated. In our study~\cite{quétu2023avoid}, it has been shown that the more challenging the dataset and classification task, the harder it is to avoid DD. In fact, we show that, for simple setups like a LeNet-300-100 trained on MNIST dataset, the sparse double descent is disappearing when the $\ell_2$ regularization is used. However, for more complex tasks, like ResNet-18 on CIFAR datasets, whether the regularization is employed or not, the sparse double descent occurs and is present in both cases. Indeed, an ablation study over $\lambda$ confirms that even for high values of $\lambda$, the phenomenon is not dodged: the overall performance of the model drops because the imposed regularization becomes too strong to allow the model to learn the training set entirely.
\begin{figure*}[t]
    \begin{subfigure}{0.49\textwidth}
        \includegraphics[width=\textwidth]{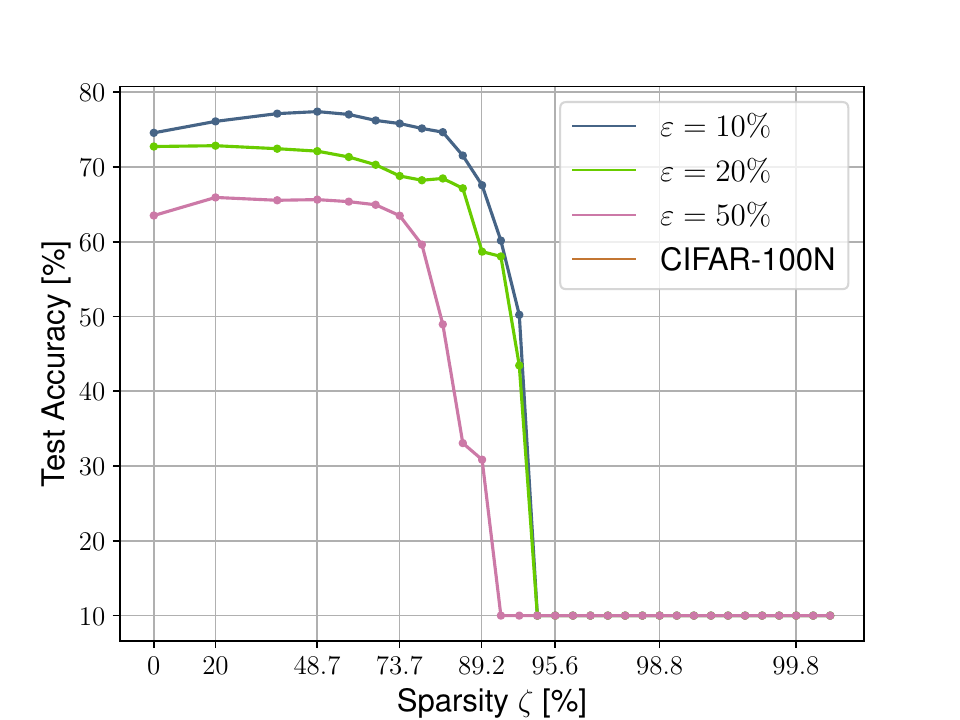}
        \caption{~}
    \end{subfigure}
    \begin{subfigure}{0.49\textwidth}
        \includegraphics[width=\textwidth]{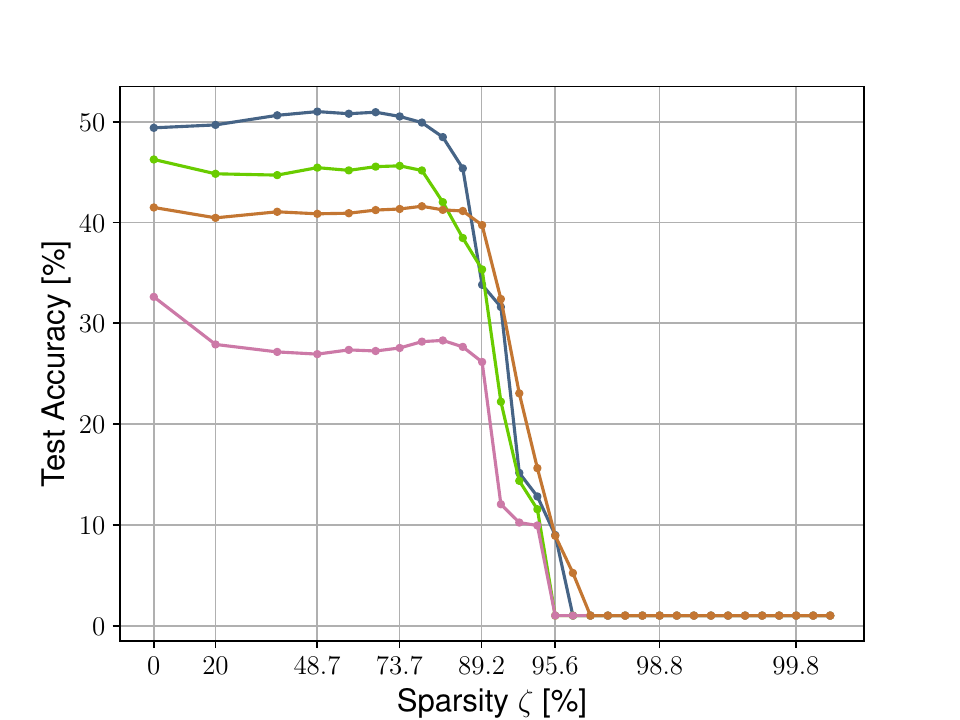}
        \caption{~}
    \end{subfigure}
    \caption{Test accuracy of ViT with different amount of noise $\varepsilon$ on CIFAR-10~(a) and CIFAR-100~(b) with $\lambda=1$ ($\varepsilon \in \{10\%, 20\%\}$), $\lambda=3$ ($\varepsilon=50\%$).}
    \label{fig:CIFAR-ViT-wd}
\end{figure*}
However, our recent work~\cite{quetu2023sparse} shows that an optimally-tuned $\ell_2$ regularization relieves sparse double descent in vision transformers (ViT). Carrying out an ablation study over $\lambda$, we demonstrate that incrementing $\lambda$ smoothens the bump of the test loss, and at some point, i.e. for $\lambda_{opt}$, the test loss becomes flat and behaves monotonically: the phenomenon is dodged. Note that for $\lambda > \lambda_{opt}$, sparse double descent is also avoided but the performance worsens as the regularization is stronger. 

\textbf{Dodging the sparse double descent.} To illustrate these facts, the results of ViT on CIFAR-10/CIFAR-100 with $\varepsilon \in \{10\%, 20\%, 50\%\}$ with $\lambda_{opt}$ are portrayed in Fig.~\ref{fig:CIFAR-ViT-wd}. With $\lambda=1$, for small noise rates, i.e. $\varepsilon \leq 20\%$, the phenomenon vanishes and performance is enhanced. However, for higher noise rates, like $\varepsilon=50\%$, the sparse double descent is mitigated, but still present. Even if it already helps, the strength of the regularization is not high enough to completely avoid the phenomenon but with a higher value of $\lambda$, \emph{i.e.,} equal to 3, the performance becomes monotonic. 

\begin{figure*}[t]
    \begin{subfigure}{0.5\textwidth}
        \includegraphics[width=\textwidth]{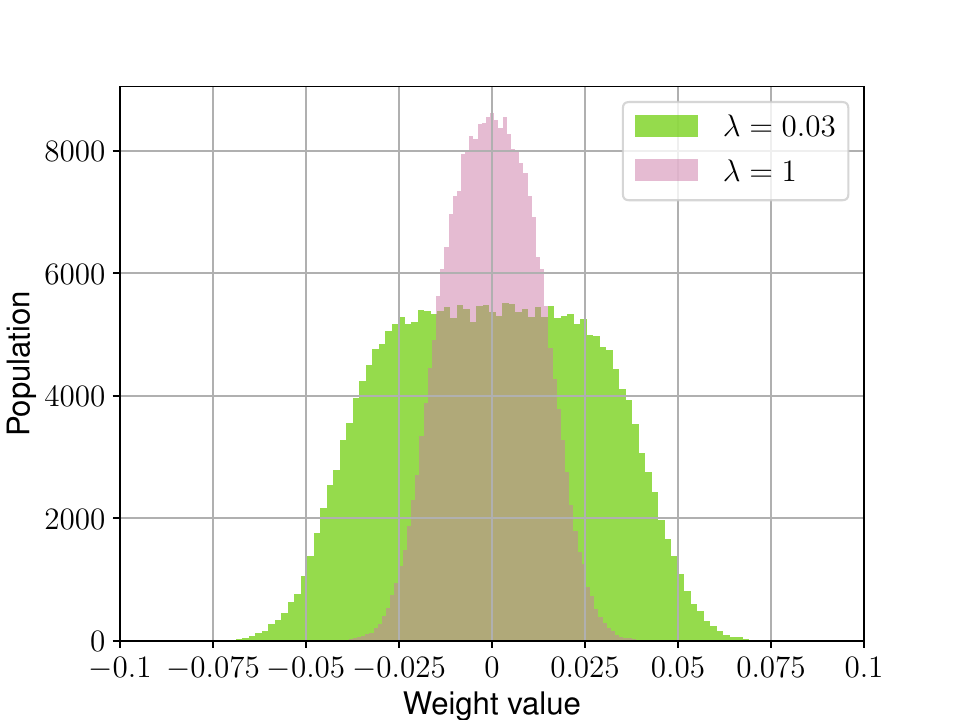}
        \caption{~}
    \end{subfigure}
    \begin{subfigure}{0.5\textwidth}
        \includegraphics[width=\textwidth]{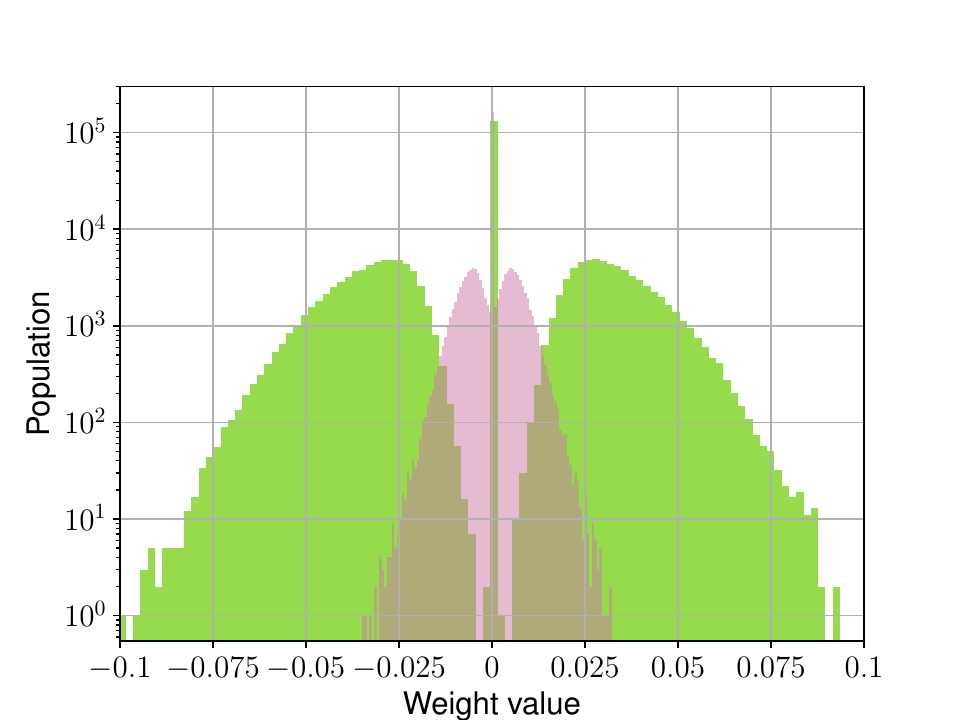}
        \caption{~}
    \end{subfigure}
    \caption{Histogram of the weights of ViT for $\varepsilon=10\%$ on CIFAR-10,\\ with $\zeta\!=\!0\%$~(a) and $\zeta=48.8\%$~(b).}
    \label{fig:Histograms}
\end{figure*}

\textbf{Performance/sparsity trade-off.}
We show in Table~\ref{tab:WD} a comparison between the use of an optimal $\ell_2$ regularization and regular techniques on CIFAR-10 with $\varepsilon=10\%$ regarding performance, sparsity and computational cost.
Vanilla training joined with traditional early stopping criteria, costing more than 289 PFLOPs (second line), leads to a model with harmed performance and lower sparsity in comparison with the entire pruning process, requiring 443 PFLOPs (first line). On the other hand, the use of the optimal $\ell_2$ regularization enabling the model to dodge sparse double descent results in a model with the lowest sparsity and worse performance than the full pruning process, but needs fewer FLOPs than the other approaches.
Indeed, as visible in Fig.~\ref{fig:CIFAR-ViT-wd}, at high sparsity \emph{i.e.,} $>90\%$, the performance is very deteriorated and even reaches random guess when $\lambda_{opt}$ is used: when we avoid SDD, the capacity of the model to be compressed suffers at high regularization regimes. This behavior is a result of the strong prior we impose on the distribution of the parameters of the model. The greater this prior, the fewer degrees of freedom we are able to eliminate from our system. The distribution of the parameters for one of the possible training setups, for $\lambda=0.03$ and $\lambda=1$, without pruning and after two pruning stages is shown in Fig.~\ref{fig:Histograms} as a visual example to illustrate this behavior.
With stronger regularization, the parameters have less variance despite being removed in the same amount, which has the dual effect of making them more robust to injected noise (due to the strong regularization), but also making this distribution more sensitive to compression by pruning. Thus, we draw the conclusion that, if we aim for a robust and well-generalizing model, we should avoid SDD and use strong $\ell_2$ regularization. On the other hand, if we aim for compressibility, we should prefer SDD because the better generalizing region is pushed to the highly compressed regions.

Although a standard regularization approach like $\ell_2$, is already positively contributing to dodging the sparse double descent, it also presents some flaws. In some setups, like ResNet-18 on CIFAR datasets, the sparse double descent is still noticeable even if $\ell_2$ regularization is used~\cite{quétu2023avoid}. Moreover, this regularization has the big drawback of sacrificing the performance/sparsity trade-off.
Thus, the need to design a custom regularization towards avoidance of sparse double descent is becoming apparent. In the next section, we present a learning scheme in which a student model is regularized by distilling knowledge from a sparse teacher in its best validation accuracy region (or even with a dense teacher), observing that the student is dodging SDD. 

\section{Distilling Knowledge Shuns Sparse Double Descent}
\label{sec:kd}
\textbf{Combining knowledge distillation and network pruning.} The combination of knowledge distillation (KD) and pruning has already been examined in various studies. The study~\cite{zhou2022efficient} proposes a ``progressive feature distribution distillation'' to improve generalization when working with a smaller training set. This method entails obtaining a student network by pruning a trained network, followed by distilling the feature distribution into the student network.
The size reduction of the final model is a primary goal of studies combining the two approaches. For instance, \cite{cui2021joint} use structured pruning and dense knowledge distillation approaches to drastically compress an initial large language model into a compressed shallow (but still deep) network.
\begin{table}[t]
\caption{Performance and training computational cost between the different approaches for ViT on CIFAR-10 with $\varepsilon=10\%$.}
    \label{tab:WD}
    \centering
    \resizebox{\textwidth}{!}{
    \begin{tabular}{c c c c c c}
    \toprule
    \bf Early & \bf \hspace{0.2cm} \multirow{2}{*} {$\boldsymbol{\lambda_{opt}}$} \hspace{0.2cm} & \bf Training cost \hspace{0.2cm} & \bf CO$_2$ emissions \hspace{0.2cm} & \bf Test accuracy \hspace{0.2cm} & \bf Sparsity  \\
     \bf stop & & \bf [PFLOPs] ($\boldsymbol{\downarrow}$) \hspace{0.2cm} & \bf [g] ($\boldsymbol{\downarrow}$) & \bf  [\%]($\boldsymbol{\uparrow}$) & \bf [\%]($\boldsymbol{\uparrow}$) \\
    \midrule
    & & 443.68 & 563.5 & 74.47 $\pm$ 0.01 & 99.26 \\
    $\checkmark$ & & 289.33 & 367.5 & 68.07 $\pm$ 0.02 & 95.60 \\
    $\checkmark$ & $\checkmark$ & 192.89 & 245.0 & 71.48 $\pm$ 0.06 & 86.58 \\
    \bottomrule
    \end{tabular}}
\end{table}
The study~\cite{kim2016sequence} revealed that standard KD applied to word-level prediction can be useful for Neural Machine Translation. Therein, on top of KD, weight pruning was used to reduce the number of parameters. Studies~\cite{chen2022knowledge} and~\cite{aghli2021combining} are examples of efforts in the area of image processing that are more practical and aimed at improving performance on a sparsified model. Moreover, the study~\cite{park2022prune} provides a number of examples where the ``prune, then distill'' approach is highly effective, drawing on the advantageous, well-generalizing qualities of both pruning and KD.

Working on a related research subject, the authors of~\cite{chang2021provable} explored whether it could be more practical to train a small model directly or rather first train a larger one and then prune it, highlighting the significance of additional research on the SDD. The authors of this work offer persuasive proof that the latter tactic is successful in improving model performance in sparsified regimes. 
The DD phenomenon in a self-supervised setting was used to give pseudo-labels to a large held-out dataset~\cite{cotter2021distilling}.
Whilst this approach attempted to exploit DD, our aim, in contrast to this work, is to avoid the sparse double descent in order to improve performance on the final student model on the same task and dataset as the teacher. Motivated by \cite{saglietti2022solvable} and building on top of \cite{park2022prune} (although in a very different context), we present in the next paragraph our approach that will drive our quest towards the dodging of SDD.

\textbf{Approach.} 
In an image classification setup, inside a KD framework, the typical objective function minimized to train a tinier student network is a linear combination of the standard cross-entropy loss $\mathcal{L}_\text{CE}$, using ``hard'' ground truth labels, and the Kullback-Leibler divergence loss $\mathcal{L}_\text{KL}$, calculated between the teacher’s predictions $\boldsymbol{y}^\text{t}$ and the student's ones $\boldsymbol{y}^\text{s}$, scaled by a temperature $\tau$ :
\begin{equation}
    \label{eq:objKL}
    \mathcal{L}=(1-\alpha)\mathcal{L}_\text{CE}(\boldsymbol{y}^\text{s},\boldsymbol{\hat{y}})+\alpha \mathcal{L}_\text{KL}(\boldsymbol{y}^\text{s},\boldsymbol{y}^\text{t}, \tau)
\end{equation}
where $\boldsymbol{\hat{y}}$ stands for the ground truth target, and $\alpha$ is the distillation hyper-parameter weighting the average between the two losses. Our KD scheme employs the same formulation loss as in~\cite{hinton2015distilling,kim2021comparing}.\\
To train and sparsify the student model, the same procedure- but changing the objective function to \eqref{eq:objKL}- as in Alg.~\ref{Algo} is employed. The teacher used to distill its knowledge can be either a pruned model in its best-fit regime (\emph{i.e.,} with the best validation accuracy) or its dense (\emph{i.e.,} unpruned) version.
\begin{figure*}[t]
    \begin{subfigure}{0.32\textwidth}
        \includegraphics[width=\textwidth]{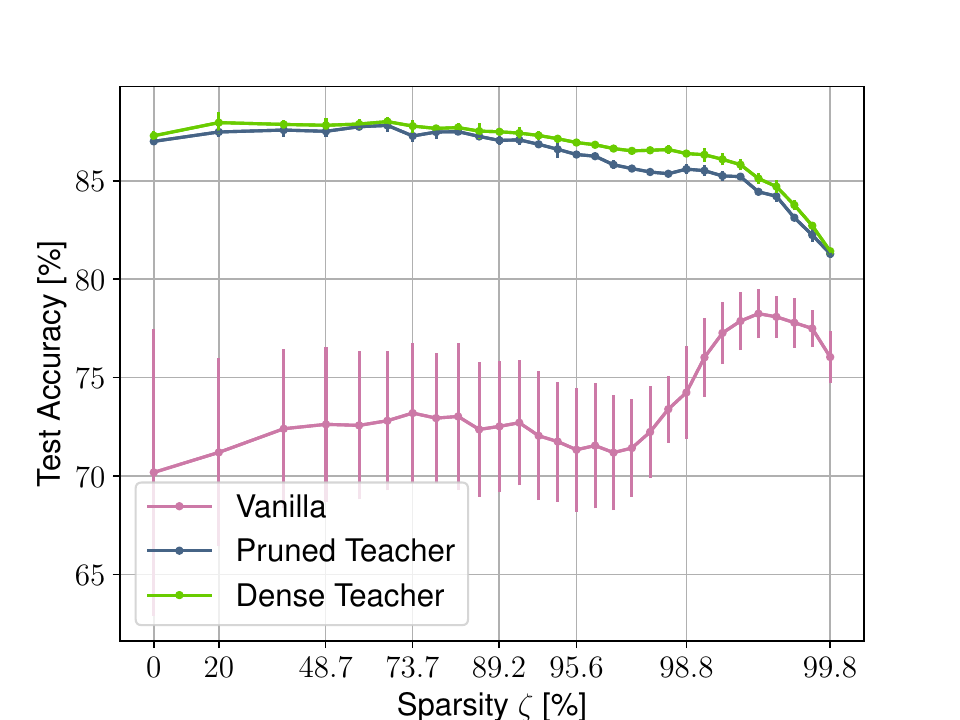}
        \caption{~}
        \label{fig:CIFAR-10_10_depth_5_width_32}
    \end{subfigure}
    \begin{subfigure}{0.32\textwidth}
        \includegraphics[width=\textwidth]{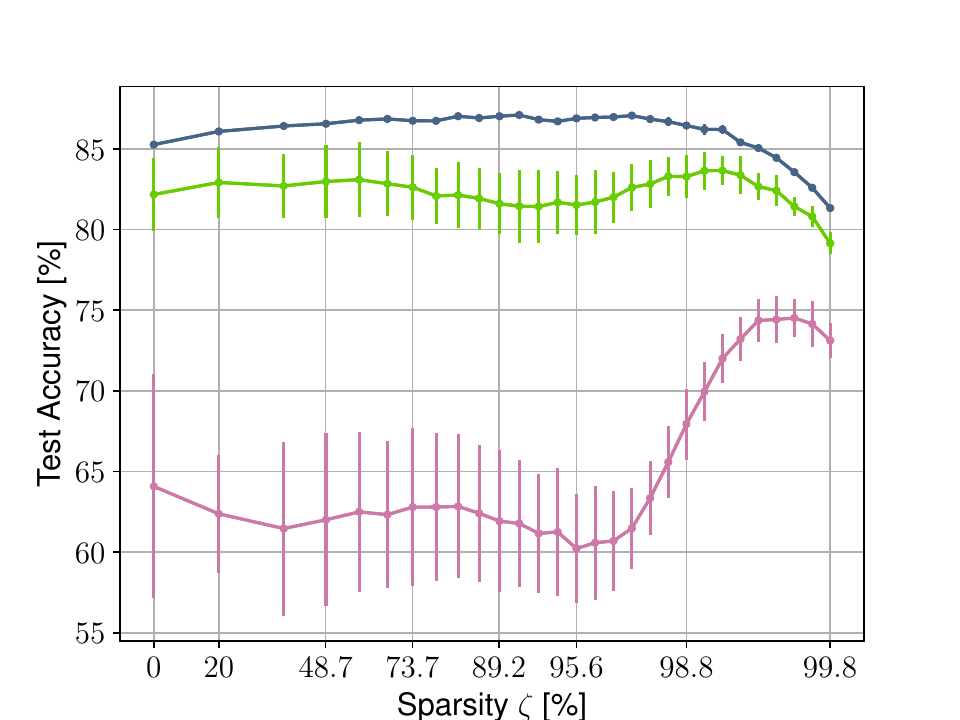}
        \caption{~}
        \label{fig:CIFAR-10_20_depth_5_width_32}
    \end{subfigure}
    \begin{subfigure}{0.32\textwidth}
        \includegraphics[width=\textwidth]{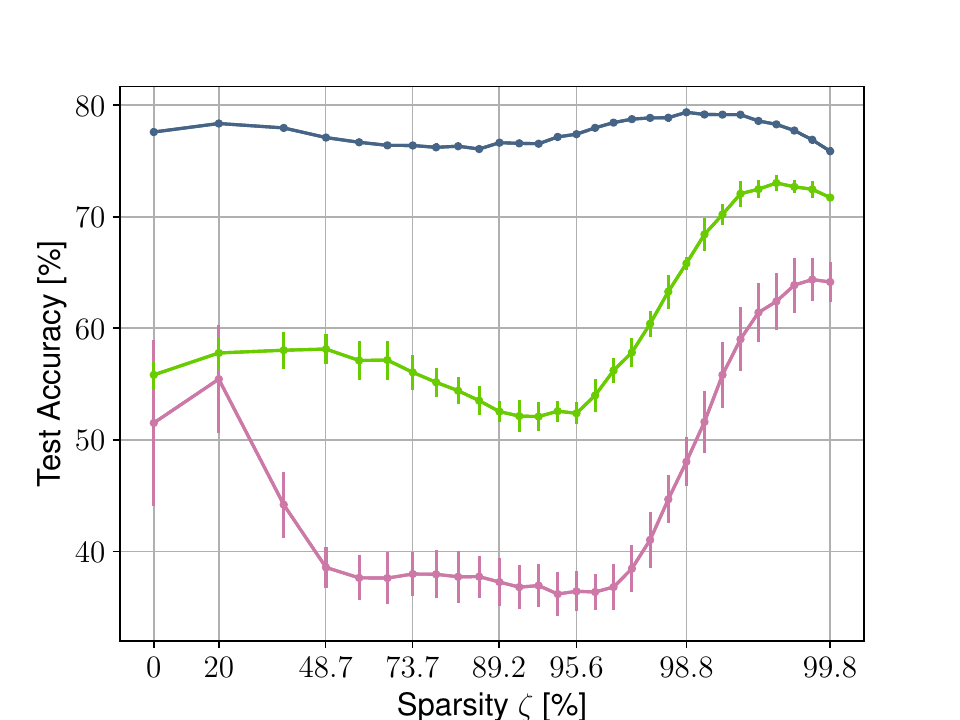}
        \caption{~}
        \label{fig:CIFAR-10_50_depth_5_width_32}
    \end{subfigure}
    \begin{subfigure}{0.32\textwidth}
        \includegraphics[width=\textwidth]{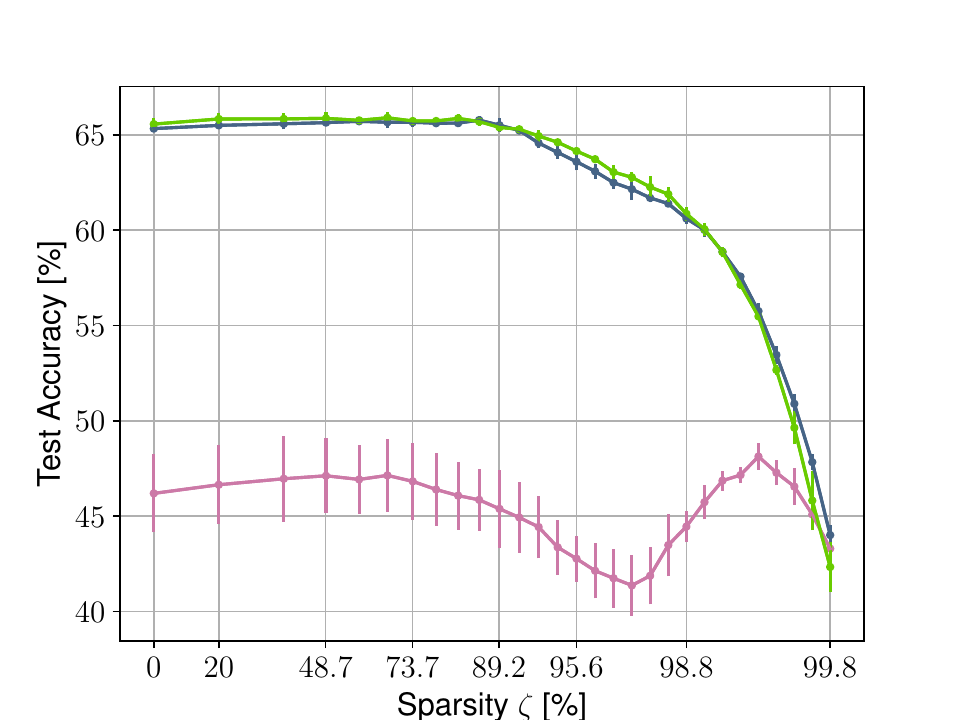}
        \caption{~}
        \label{fig:CIFAR-100_10_depth_5_width_32}
    \end{subfigure}
    \begin{subfigure}{0.32\textwidth}
        \includegraphics[width=\textwidth]{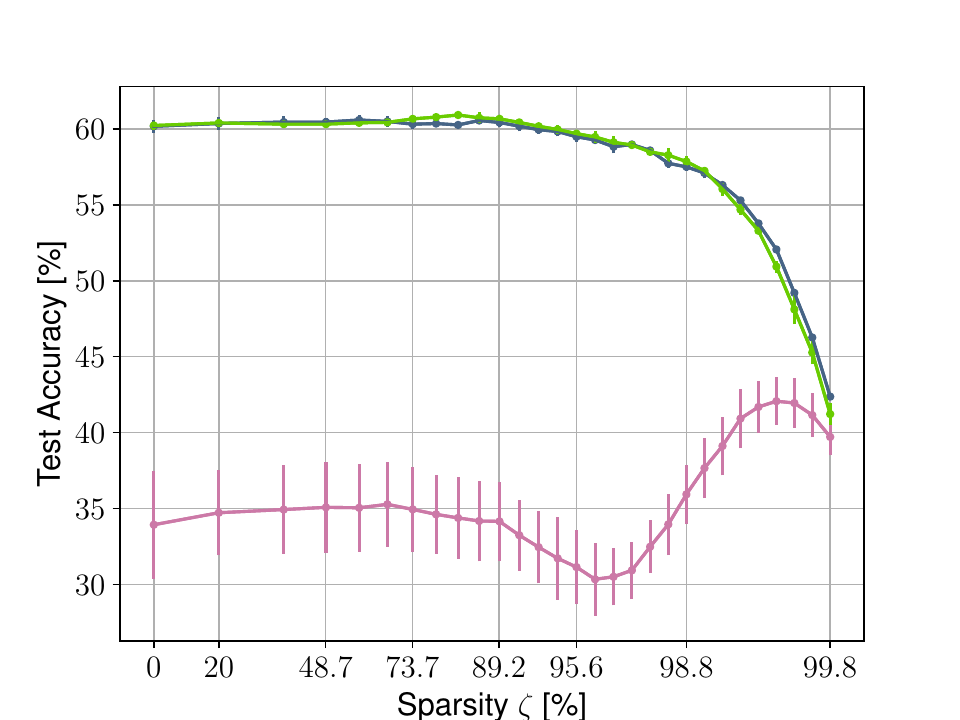}
        \caption{~}
        \label{fig:CIFAR-100_20_depth_5_width_32}
    \end{subfigure}
    \begin{subfigure}{0.32\textwidth}
        \includegraphics[width=\textwidth]{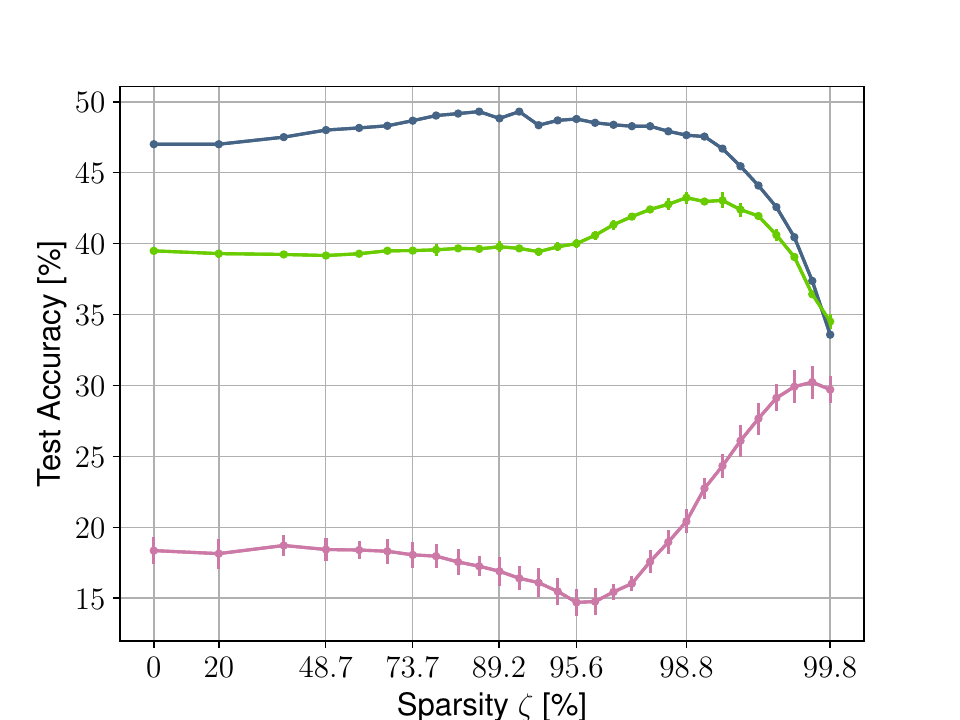}
        \caption{~}
        \label{fig:CIFAR-100_50_depth_5_width_32}
    \end{subfigure}
    \caption{Performance of the VGG-like model on CIFAR-10~(a,b,c) and CIFAR-100~(d,e,f) for different label noises. \textbf{Left.} $\varepsilon=10\%$ \textbf{Middle.} $\varepsilon=20\%$ \textbf{Right.} $\varepsilon=50\%$}
    \label{fig:mainres}
\end{figure*}

\textbf{Results.} 
To conduct the following experiments, a ``VGG-like'' model is defined characterized by its depth $\delta$ (the larger, the deeper the model) and the number of convolutional filters per layer $2^{\gamma}$ (the larger, the wider the model) following~\cite{quetu2023dodging}.
The results on CIFAR-10 and CIFAR-100 with $\varepsilon \in \left\{10\%, 20\%, 50\%\right\}$ are portrayed in Fig.~\ref{fig:mainres}. A ResNet-18 teacher, whose performance can be found in Fig.~\ref{fig:CIFAR-ViT-ResNet}, is used to distill the knowledge in a VGG-like student model with $\gamma=32$ and $\delta=5$.
For every value of $\varepsilon$, the sparse double descent phenomenon is always revealed by the student model trained with a vanilla-training setup. Nonetheless, the same student model trained within our KD scheme is consistently eschewing this phenomenon, whether the knowledge is distilled from the pruned or the dense teacher.   


\textbf{Computation/performance trade-off.} 
The comparison between our approach and traditional techniques on CIFAR-10 with $\varepsilon=20\%$ regarding performance achieved and computational cost at training time is presented in Table~\ref{tab:Computational Cost CIFAR-10 20}.
Vanilla training joined with traditional early stopping criteria, (second line) leads to a model with low sparsity and poor performance. To reach decent performance, the whole pruning process has to be entirely completed until all of its parameters are completely removed, which costs more than 48 PFLOPs (first line).
On the other hand, our method produces a model with high sparsity achieving more than 10\% improvement in performance with approximately 25\% less computation.
We believe this to be a core practical contribution in real-world scenarios, particularly when working with limited and noisy annotated datasets, where SDD can easily occur.

\begin{table}[t]
\caption{Performance and training computational cost between traditional approaches and our scheme for a VGG-like model on CIFAR-10 with $\varepsilon=20\%$.}
    \label{tab:Computational Cost CIFAR-10 20}
    \centering
    \resizebox{\textwidth}{!}{
    \begin{tabular}{c c c c c c c}
    \toprule
    \bf Early & \bf \multirow{2}{*}{Distillation} & \bf Distillation from & \bf Training FLOPs & \bf CO$_2$ & \bf Test accuracy& \bf Sparsity \\
    \bf stop & & \bf pruned teacher & \bf [PFLOPs] ($\boldsymbol{\downarrow}$) & \bf emissions [g] ($\boldsymbol{\downarrow}$) & \bf  [\%]($\boldsymbol{\uparrow}$) & \bf [\%]($\boldsymbol{\uparrow}$) \\
    \midrule
    & & & 48.84 & 510 & 74.52 $\pm$ 1.20 & 99.62 \\
    $\checkmark$ & & & 4.88 & 51 & 60.23 $\pm$ 3.37 & 36.00 \\ 
    $\checkmark$ & $\checkmark$ & & 35.82 & 374 & 81.52 $\pm$ 1.85 & 99.26 \\
    $\checkmark$ & $\checkmark$ & $\checkmark$ & 35.82 & 374 & 86.89 $\pm$ 0.16 & 99.26 \\
    \bottomrule
    \end{tabular}}
\end{table}

\section{Environmental Impact}
\label{sec:Environmental}

\textbf{Carbon emissions at training time.} Besides comparing the training cost, in terms of FLOPs required to achieve the models with the different mentioned approaches above, we also present the estimation of emitted CO$_2$ at training time using~\cite{anthony2020carbontracker} in Tables~\ref{tab:WD} and~\ref{tab:Computational Cost CIFAR-10 20}.

Despite leading to a model with a worse performance/sparsity trade-off, the use of an optimal $\ell_2$ regularization is the greener approach, as it can be observed in Table~\ref{tab:WD}, saving at least 30\% of carbon emissions compared to the two other methods. Indeed, the vanilla approach without early stopping criteria, is offering the best performance/sparsity compromise but is the biggest emitter of CO$_2$ with more than 550g of carbon emissions.

Concerning our KD scheme, vanilla training joined with traditional early stopping criteria is the greener approach, emitting 51g of CO$_2$, but has the big drawback to lead to a model with low sparsity and poor performance, as given in Table~\ref{tab:Computational Cost CIFAR-10 20}. Since the whole pruning process has to be entirely completed until all of its parameters are completely removed to achieve better performance, the vanilla approach is the biggest polluter with more than 500g of carbon emissions. 
On the other hand, our method, producing a model with high sparsity and achieving more than 10\% improvement in performance, saves more than 25\% of CO$_2$ emissions.

\textbf{CO$_2$ at inference time.} Our presented approaches already enable us to reduce the carbon footprint at training time. However, since a neural network is going to be used multiple times for inference, it is also important to lessen the CO$_2$ emissions related to this use. Indeed, the chosen unstructured pruning technique is only limited to the identification of the portion of the parameters that can be set to zero. Therefore, it offers very few, if any, practical benefits when it comes to the deployment of the model in a resource-constrained system. 
To make the most of it in real-life scenarios and address this problem, a study to optimally deploy our models onto an edge device should be carried on, and the use of specifically designed libraries like~\cite{bragagnolo2021simplify} can be of great help to meet this type of need. We leave this aspect for future work.

\section{Conclusion}
\label{sec:Conclusion}

In this paper, we have proposed two approaches to shun the sparse double descent phenomenon: a simple $\ell_2$ regularization and a learning scheme involving knowledge distillation. 
While standard regularization methods like $\ell_2$ are positively contributing to lessening such a phenomenon, they are revealing some limits, which can be resolved by leveraging a knowledge distillation scheme. Thanks to that scheme, the good generalization properties of the teacher are transmitted to the student model, which is no longer suffering from sparse double descent. 

Sparse double descent is impacting the determination of the optimal model size required to maintain the performance of over-parametrized models. With this study, we hope to inform the community about this risk that various deep neural networks might face in compression schemes.

\newpage
\bibliographystyle{splncs04}
\bibliography{main}

\end{document}